\documentclass[conference]{IEEEtran}
\IEEEoverridecommandlockouts
\usepackage{amsmath,amssymb,amsfonts}
\usepackage{algorithmic}
\usepackage{graphicx}
\usepackage{textcomp}
\usepackage{xcolor}
\usepackage{hyperref}

\hypersetup{
    colorlinks=true,
    linkcolor=black,
    filecolor=black,      
    urlcolor=black,
    citecolor=black
    }

\usepackage[numbers,square]{natbib}
\def\BibTeX{{\rm B\kern-.05em{\sc i\kern-.025em b}\kern-.08em
    T\kern-.1667em\lower.7ex\hbox{E}\kern-.125emX}}
\begin{document}

\title{RMDM: A Multilabel Fakenews Dataset for Vietnamese Evidence Verification}

\author{
Hai-Long Nguyen$^1$, Thi-Kieu-Trang Pham$^1$, Thai-Son Le$^1$, Tan-Minh Nguyen$^1$,\\ Thi-Hai-Yen Vuong$^{1}$, Ha-Thanh Nguyen$^2$ \\ 
\textit{$^1$VNU University of Engineering and Technology}, Hanoi, Vietnam \\ 
\textit{$^2$National Institute of Informatics}, Tokyo, Japan \\ 
\{$^1$long.nh, 20020485, 20020069, 20020081, yenvth\}@vnu.edu.vn \\
$^2$nguyenhathanh@nii.ac.jp
}


\maketitle

\begin{abstract}
In this study, we present a novel and challenging multilabel Vietnamese dataset (RMDM) designed to assess the performance of large language models (LLMs), in verifying electronic information related to legal contexts, focusing on fake news as potential input for electronic evidence. The RMDM dataset comprises four labels: real, mis, dis, and mal, representing real information, misinformation, disinformation, and mal-information, respectively. By including these diverse labels, RMDM captures the complexities of differing fake news categories and offers insights into the abilities of different language models to handle various types of information that could be part of electronic evidence. The dataset consists of a total of 1,556 samples, with 389 samples for each label. Preliminary tests on the dataset using GPT-based and BERT-based models reveal variations in the models' performance across different labels, indicating that the dataset effectively challenges the ability of various language models to verify the authenticity of such information. Our findings suggest that verifying electronic information related to legal contexts, including fake news, remains a difficult problem for language models, warranting further attention from the research community to advance toward more reliable AI models for potential legal applications.
\end{abstract}

\section{Introduction}

The Civil Procedure Code of 2015 (CPC) plays a significant role in addressing civil and commercial cases in practice. One of the progressive regulations in the CPC is the introduction of ``electronic data'' as a source of evidence, commonly referred to as ``electronic evidence'', following the 2006 Electronic Transactions Law (ETL). According to the ETL, ``data'' refers to information in the form of symbols, writings, numbers, images, sound, or similar formats, revealing that ``electronic data'' is regarded as a source of evidence in electronic transactions.

Large Language Models (LLMs) \cite{vaswani2017attention}, such as GPTs \cite{radford2019language,brown2020language,openai2023gpt4}, have contributed significantly to advances in natural language processing tasks, including text classification, sentiment analysis, and content generation, among others. However, the potential of LLMs in the context of verifying the authenticity of electronic evidence in legal proceedings remains underexplored \cite{thanh2021summary,nguyen2022attentive, nguyen2022transformer, vuong2022sm}. Therefore, we propose a novel dataset tailored to evaluating LLMs' performance in verifying electronic evidence.

Our dataset, RMDM, is designed to assess the capabilities of LLMs in verifying electronic evidence in Vietnamese legal proceedings. The dataset includes four labels: real, mis, dis, and mal, which represent real information, misinformation, disinformation, and mal-information as electronic evidence, respectively.

By introducing a diverse set of labels for electronic evidence, our RMDM dataset enables a more granular analysis of various types of fake news. This multilabel approach captures the complexities of differing fake news categories, which can inform the development and evaluation of LLMs that are better equipped to handle the varied nature of electronic evidence in legal proceedings.

The main contributions of this paper are as follows:

\begin{itemize}
    \item Development of a new dataset with four labels suitable for evaluating various language models, such as GPTs and BERT, in the context of verifying electronic evidence in legal proceedings, capturing the complexities of differing fake news categories and offering insights into their abilities to handle various types of electronic evidence.
    \item Assessment of the electronic evidence verification capabilities of state-of-the-art generative models, including ChatGPT (GPT-3.5) and GPT-4, providing an empirical evaluation of their strengths and weaknesses in detecting different fake news types.
    \item Identification of key challenges associated with language models' performance in verifying electronic evidence, highlighting the need for further research and development in AI models for legal applications and paving the way for potential integration of external knowledge bases and logical verifiers to enhance their understanding of information and context.
\end{itemize}

\section{Background}

The digital era has brought increasing importance to electronic evidence in legal proceedings. Electronic evidence can encompass various forms of electronic data, including both real and fake information. Fake news presents a significant challenge in legal contexts, as it can be intentionally deceptive and difficult to differentiate from legitimate electronic evidence. Given the prevalence of fake news on social media and other online platforms, the need for reliable methods to verify electronic evidence, particularly in distinguishing between real information and fake news, is critical for ensuring accurate and fair decision-making in legal proceedings.

\subsection{Definition of Fake News}
Although there are many studies related to fake news, there is no common and official definition for fake news. One of the first definitions, widely used in many studies, was provided by Shu et al. (2017) \cite{shu2017fake}, which defined fake news as \textbf{``fake news is a news article that is intentionally and verifiably false.''} This definition limits the types of information in fake news, intending to deceive users and disregarding some misleading information. According to Siva Vaidhyanathan - Professor of Communication at the University of Virginia, author of ``Antisocial Media: How Facebook Disconnects Us and Undermines Democracy,'' fake news is a piece of news or information shared with the intent to confuse, deceive, or undermine the trust of some people in the public. It can be distorted information, inaccurate information, or entirely fabricated stories and is often spread quickly on the network by users' sharing \cite{vaidhyanathan2018antisocial}. However, according to Claire Wardle, co-founder and leader of First Draft \footnote{https://firstdraftnews.org/}, the term fake news is not enough to describe the complexity of different types of information phenomena. Instead, she divided it into three types of information: Misinformation, Disinformation, and Mal-information \footnote{https://money.cnn.com/2017/11/03/media/claire-wardle-fake-news-reliable-sources-podcast/index.html}.

\subsection{Classification of Fake News}
Fake news exists in various forms, so classifying fake news is not an easy task. There are many studies on the classification of fake news. The authors introduce the classification of fake news according to Claire Wardle \cite{wardle2017information}. Accordingly, fake news is divided into the following three categories:

\begin{enumerate}
    \item Misinformation~\cite{west2021misinformation,swire2020public,jerit2020political}: Misinformation is created but not intended to harm, including the following two types:
    \begin{itemize}
        \item False connection: When the headline, image, or caption does not match the content. For example, information that is not entirely false can be distorted using misleading or sensational headlines.
        \item Misleading content: Using incorrect information and causing misunderstanding for the reader. For example, the statement ``Scientists claim that eating chocolate every day can help you lose weight'' can be misleading to readers. Some studies have suggested that chocolate may have health benefits, such as improving cardiovascular health and reducing stress, but there is no scientific evidence to support the claim that eating chocolate every day can help you lose weight.
    \end{itemize}
    \item Disinformation~\cite{fallis2014varieties,fallis2015disinformation}: Inaccurate information, created and shared by those who intend to harm, includes the following types:
    \begin{itemize}
        \item False context: Verified information is shared with false information about the context (adjusted in dangerous ways). For example, a video circulated on social media showing the screams of a female student in school, with the introduction that the screams were due to harassment. However, it was just the screams of a female student who could not control her emotions in an argument.
        \item Imposter content: False information is created and circulated to impersonate someone or an organization by using their famous features. This form is often used to deceive, steal information, and affect the reputation and credibility of the person being impersonated. A common example is impersonating bank employees or police officers to call the victims and get their personal information to seize assets in bank accounts or savings.
        \item Manipulated content: When some aspects of the original content are changed. This usually involves images or videos. For example, when intentionally spreading false information about a celebrity scandal, a commonly used trick is photo editing or modifying the original image to illustrate the content.
        \item Fabricated content: 100\% false content designed to fabricate and cause harm. This type of fabricated content is often created and spread with the goal of attracting attention or generating clicks, rather than providing accurate and truthful information. For example, ``Tran Thanh and many other Vietnamese stars have embezzled charity money for the Central region'' is fabricated content and has no basis in reality, as there is no evidence or foundation to verify the accuracy of this information.
    \end{itemize}
    \item Mal-information~\cite{wardle2018thinking,baines2020defining}: True information that has been reconfigured or manipulated in a way that is misleading or harmful. It includes:
    \begin{itemize}
        \item Leaks: Leaking information is an event that occurs when secret information is disclosed to unauthorized persons or parties. For example, in US presidential elections or before Party congresses in Vietnam, there are often many pieces of information believed to be leaked from confidential documents and almost impossible to verify. This information often causes confusion and creates many conflicting public opinions.
        \item Harassment: Any behavior, whether through speech, images, text, or otherwise, intended to offend or humiliate an individual or organization. With social media, harassment is becoming more prevalent and sophisticated. For example, fan pages of celebrities often spread information to discredit competitors and raise their idol's image.
        \item Hate speech: Content expressed through speech, text, or other expressions that show hatred or ridicule towards a person or other people. Hate speech often targets a social group defined by attributes such as race, ethnicity, gender, or sexual orientation.
    \end{itemize}
\end{enumerate}

\subsection{Overview of Some Vietnamese Fake News Datasets}

In this section, we briefly discuss a few existing Vietnamese fake news datasets as a foundation and reference for developing our dataset:

\begin{itemize}
    \item VFND \cite{vfnd}: Vietnamese Fake News Dataset. This dataset includes 2,070 news articles collected between 2017 and 2019, comprising 1,035 fake news articles and 1,035 real ones. The news articles were classified as real or fake based on various sources, cross-referencing the cited sources, or community-based classification.
    \item VNTC \cite{hoang2007comparative}: A Large-scale Vietnamese News Text Classification Corpus. This dataset encompasses 2,010 news articles with 1,005 fake news articles and 1,005 real ones, gathered from various sources, including social media and online forums, covering an array of topics such as politics, health, and entertainment.
    \item VLSP Fake News: Part of a shared task organized by the Vietnamese Language and Speech Processing group over several years, this dataset contains both fake and real news articles in Vietnamese. It focuses on political news gathered from various sources, including social media, and has been manually labeled. The dataset consists of 8,170 news articles, including 2,040 fake news articles and 6,130 real ones.
\end{itemize}

A key limitation of existing Vietnamese fake news datasets, such as VFND, VNTC, and VLSP Fake News, is that they predominantly focus on binary classification, which may oversimplify the complexity of fake news types. To address this, our RMDM dataset introduces four distinct labels - real, misinformation, disinformation, and mal-information - representing a diverse range of information and intent. This more comprehensive classification enables a deeper understanding of various fake news categories and promotes the development of robust models suitable for broader applications, particularly in legal contexts where verifying electronic information accuracy and credibility is crucial. For instance, having four distinct labels allows for better identification of electronic evidence that may have been manipulated with malicious intent, as well as evidence that may simply contain unintentional errors, thus providing a more accurate basis for decision-making in legal proceedings.

\section{Dataset}

\subsection{Dataset Overview}
The dataset, referred to as RMDM, consists of 1556 samples, with each sample containing two fields: \textit{text} and \textit{label}. The details of these fields are as follows:
\begin{itemize}
    \item \textit{text}: a short snippet containing news information.
    \item \textit{label}: the label of the sample, which belongs to one of the following four categories: real, mis, dis, and mal.
\end{itemize}
The label field comprises: 1) \textit{real}: representing true or legitimate news sources; 2) \textit{mis}: denoting misleading information that causes confusion for readers; 3) \textit{dis}: indicating fabricated and deliberately harmful news; and 4) \textit{mal}: referring to leaked information that deliberately causes harm. The dataset includes 389 samples for each label.

\subsection{Text Field Description}

\begin{figure}
    \centering
    \includegraphics[width=0.8\linewidth]{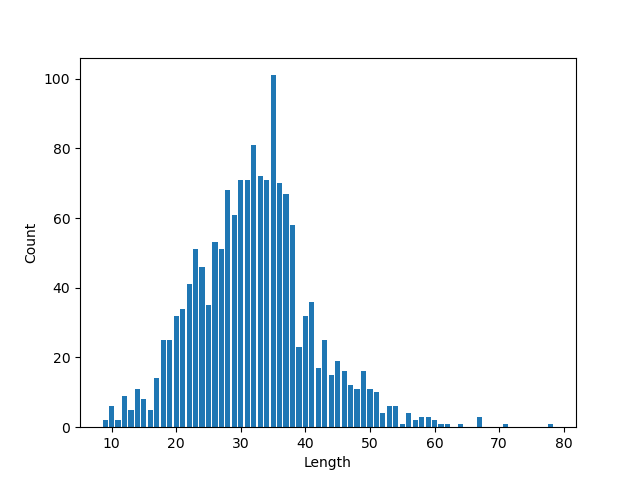}
    \caption{Distribution of word count in the text field across the entire dataset. The horizontal axis represents the number of words, while the vertical axis represents the number of documents.}
    \label{fig:word_count_whole}
\end{figure}

The average word count in the text field is 31, with the longest sentence having 78 words and the shortest having 9. The distribution of word count in the text field for the entire dataset with varying lengths is illustrated in Figure \ref{fig:word_count_whole}.

\subsubsection{Real Label}
\begin{figure}
    \centering
    \includegraphics[width=0.8\linewidth]{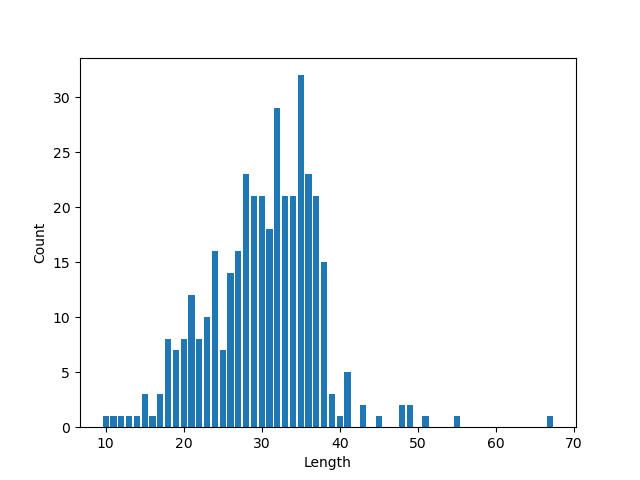}
    \caption{Distribution of word count in the text field for the real label. The horizontal axis represents the number of words, while the vertical axis represents the number of documents.}
    \label{fig:word_count_real}
\end{figure}

The average word count in the text field for the real label is 30, with the longest sentence having 67 words and the shortest having 10. The distribution of word count in the text field for the real label with varying lengths is illustrated in Figure \ref{fig:word_count_real}.

\subsubsection{Mis Label}
\begin{figure}
    \centering
    \includegraphics[width=0.8\linewidth]{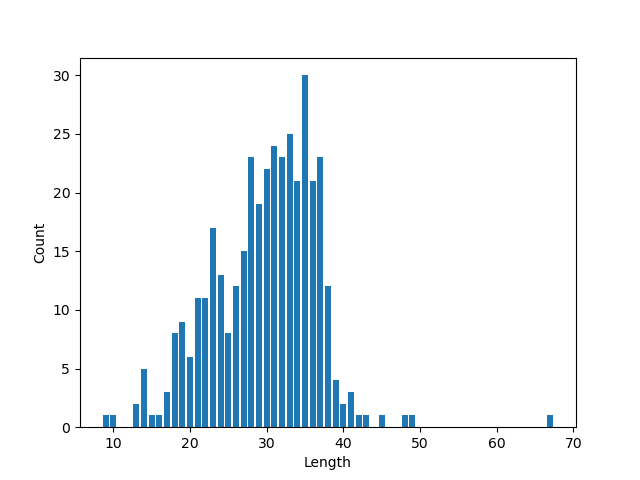}
    \caption{Distribution of word count in the text field for the mis label. The horizontal axis represents the number of words, while the vertical axis represents the number of documents.}
    \label{fig:word_count_mis}
\end{figure}

The average word count in the text field for the mis label is 30, with the longest sentence having 67 words and the shortest having 9. The distribution of word count in the text field for the mis label with varying lengths is illustrated in Figure \ref{fig:word_count_mis}.

\subsubsection{Dis Label}
\begin{figure}
    \centering
    \includegraphics[width=0.8\linewidth]{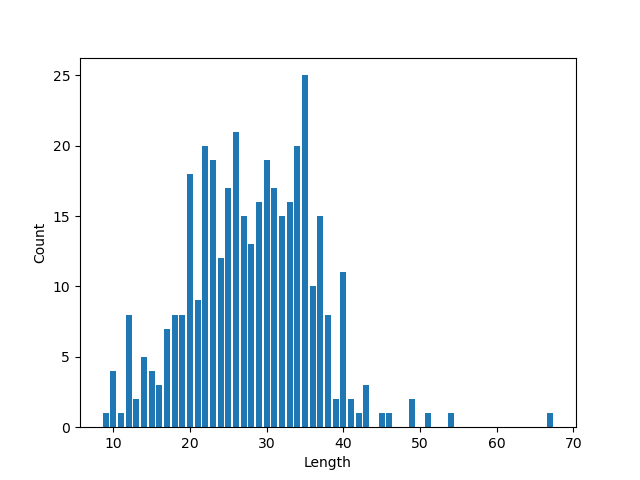}
    \caption{Distribution of word count in the text field for the dis label. The horizontal axis represents the number of words, while the vertical axis represents the number of documents.}
    \label{fig:word_count_dis}
\end{figure}

The average word count in the text field for the dis label is 28, with the longest sentence having 67 words and the shortest having 9. The distribution of word count in the text field for the dis label with varying lengths is illustrated in Figure \ref{fig:word_count_dis}.

\subsubsection{Mal Label}
\begin{figure}
    \centering
    \includegraphics[width=0.8\linewidth]{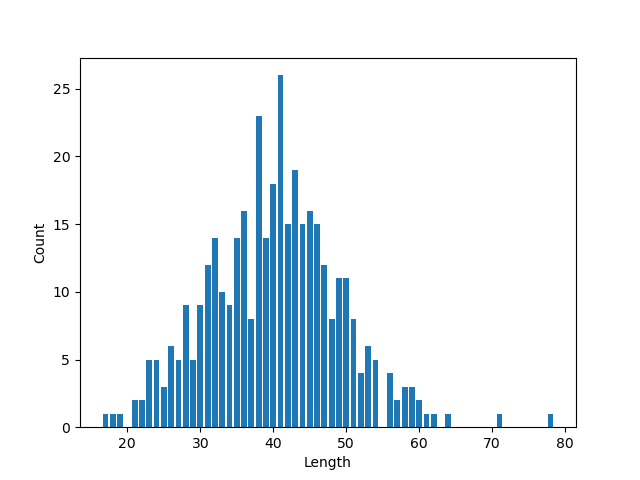}
    \caption{Distribution of word count in the text field for the mal label. The horizontal axis represents the number of words, while the vertical axis represents the number of documents.}
    \label{fig:word_count_mal}
\end{figure}

The average word count in the text field for the mal label is 40, with the longest sentence having 78 words and the shortest having 9. The distribution of word count in the text field for the mal label with varying lengths is illustrated in Figure \ref{fig:word_count_mal}.

The four-label structure of the RMDM dataset allows for a more comprehensive understanding of the characteristics, nuances, and interrelations of different types of electronic evidence. By incorporating this diverse set of labels, we aim to simulate the complexity and variety that LLMs may encounter in real-world legal scenarios, which could contribute to a more accurate and robust verification of electronic evidence.

\subsection{Label Field Description}
We assigned 388 samples for each label. The false news types with labels mis, dis, and mal were created by altering the true news - news with the real label. The team built the dataset using the following steps:

First, true news snippets, labeled as real, were extracted from reputable sources, specifically from Vietnamnet and VTV websites. The news information was selected from various domains, including domestic and international news, business news, sports news, healthcare news, and technology news. The time frame of the snippets mainly falls within the years 2019 and 2020.

Next, for each of the three types of false news, we created false news from true news. For samples with the mis label, the text field was populated by slightly altering some information (but not the entire content) in the samples with the real label. The team changed information that was deemed to have little impact on the content, such as dates, locations, etc. This resulted in samples that contain both types of misleading information - causing confusion and incorrect connections. For samples with the dis label, the text field was created by changing almost all the information or turning it into a negation of the samples with the real label. These samples with the dis label mainly fall into the categories of fabricated and counterfeit information. For samples with the mal label, we added some judgmental or sarcastic sentences to the samples with the real label, thus primarily creating divisive information.

\section{Experiment}

\subsection{Experiment Setup}


To construct the baseline model for the dataset, state-of-the-art models such as chatGPT (GPT 3.5) and GPT-4, or the model with classic BERT architecture \cite{DBLP:journals/corr/abs-1810-04805} and pre-trained parameters Multilingual-BERT \footnote{\url{https://huggingface.co/bert-base-multilingual-cased}} have been employed. The objective of these experiments is to assess the effectiveness of the most prevalent language models on the task of determining how well these state-of-the-art models can verify electronic evidence in the context of Vietnamese legal proceedings. 

Evaluating the performance of advanced language models on the RMDM dataset, which consists of four distinct labels, sheds light on their success and shortcomings in dealing with different types of electronic evidence. This multilabel evaluation indicates the adaptability and reliability of LLMs in verifying electronic evidence and understanding the complexities of fake news categories, demonstrating the value of four-label datasets in advancing AI applications in the legal domain.

\subsection{Results and Analysis}

\begin{figure}
    \centering
    \includegraphics[width=0.8\linewidth]{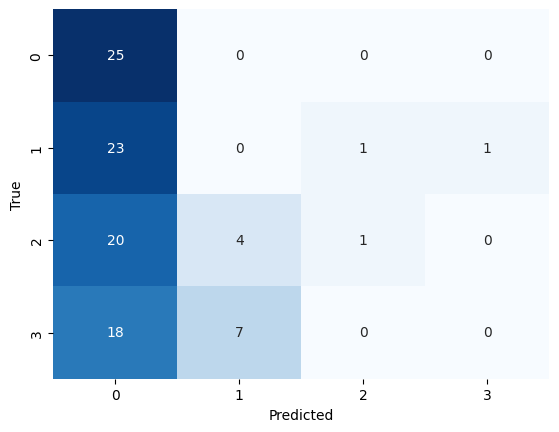}
    \caption{Confusion matrix of ChatGPT (GPT-3.5) in percentage format.}
    \label{fig:chatgpt_confusion_matrix}
\end{figure}

\begin{figure}
    \centering
    \includegraphics[width=0.8\linewidth]{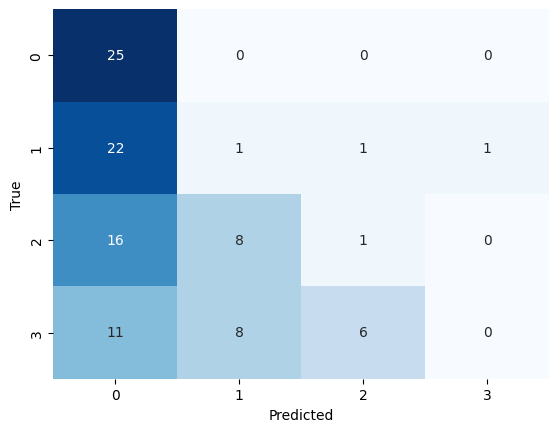}
    \caption{Confusion matrix of GPT-4 in percentage format.}
    \label{fig:gpt4_confusion_matrix}
\end{figure}

\begin{figure}
    \centering
    \includegraphics[width=0.8\linewidth]{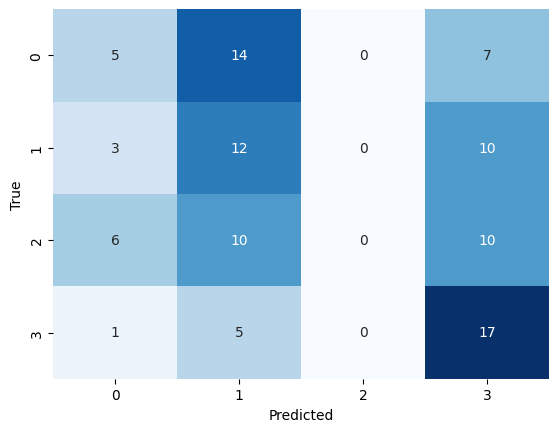}
    \caption{Confusion matrix of Multilingual-BERT based in percentage format.}
    \label{fig:multilingual_bert_confusion_matrix}
\end{figure}

Figures \ref{fig:chatgpt_confusion_matrix}, \ref{fig:gpt4_confusion_matrix}, and \ref{fig:multilingual_bert_confusion_matrix}  illustrate the confusion matrices for ChatGPT, GPT-4 models and BERT-based model with pre-trained Multilingual BERT respectively, presented in percentage format.

The revised results and analysis based on the actual confusion matrices demonstrate that all models indeed struggle to accurately predict fake news labels, with a strong bias towards classifying most information as real news.

For ChatGPT:
\begin{itemize}
    \item 25\% of samples were correctly labelled as real news.
    \item 23\% of samples were misclassified from misinformation (mis) to real news.
    \item 20\% of samples were misclassified from disinformation (dis) to real news.
    \item 18\% of samples were misclassified from mal-information (mal) to real news.
\end{itemize}

For GPT-4:
\begin{itemize}
    \item 25\% of samples were correctly labelled as real news.
    \item 23\% of samples were misclassified from misinformation (mis) to real news.
    \item 16\% of samples were misclassified from disinformation (dis) to real news.
    \item 11\% of samples were misclassified from mal-information (mal) to real news.
\end{itemize}

For BERT-based model with Multilingual-BERT parameters:
\begin{itemize}
    \item 5\% of samples were correctly labelled as real news.
    \item 3\% of samples were misclassified from misinformation (mis) to real news.
    \item 6\% of samples were misclassified from disinformation (dis) to real news.
    \item 1\% of samples were misclassified from mal-information (mal) to real news.
\end{itemize}

The models performed poorly in differentiating between mis, dis, and mal news types. This consistent bias towards real news reveals the models' lack of a proper understanding of fake news and highlights the difficulty in effectively detecting various fake news types.

The observed limitations of these advanced language models in verifying electronic evidence raise critical challenges for their application in the field of AI and Law. To overcome these challenges, one potential solution is to incorporate external knowledge bases and logical verifiers into the LLMs \cite{satoh2011proleg,nguyen2022multi}. Such integration could provide models with additional context and factual information, enhancing their ability to verify the authenticity of the input data.

As LLMs, such as ChatGPT and GPT-4, become increasingly popular in various fields, addressing these challenges with comprehensive datasets and improved solutions becomes essential. Further investigation into methods and techniques to improve electronic evidence verification capabilities is crucial to responsibly harness AI and law applications.

\section{Conclusions}
In conclusion, the RMDM dataset evaluates the performance of language models, such as GPTs and BERT, in verifying electronic evidence in Vietnamese legal proceedings. This multilabel dataset captures the complexities of fake news categories and highlights the limitations of state-of-the-art models in this context. The findings emphasize the need for further research and improvements in AI models for legal applications, and the potential integration of external knowledge bases and logical verifiers to enhance language models' understanding of information and context.

\section*{Acknowledgement}
Hai-Long Nguyen was funded by the Master, PhD Scholarship Programme of Vingroup Innovation Foundation (VINIF), code VINIF.2022.ThS.050.

\bibliographystyle{abbrvnat}
\bibliography{ref}

\end{document}